\documentclass[times,twocolumn,final,nopreprintline]{elsarticle}
\usepackage{geometry}
\usepackage{framed,multirow}

\usepackage{amssymb}
\usepackage{latexsym}

\usepackage{url}
\usepackage{xcolor}
\usepackage{makecell}
\usepackage{tabu, booktabs}
\usepackage{multirow, multicol}
\usepackage{amsmath,amssymb,amsfonts}
\usepackage{comment}
\usepackage{hyperref}
\usepackage{bigstrut}
\usepackage{enumitem}
\usepackage{adjustbox}
\definecolor{newcolor}{rgb}{.8,.349,.1}
\usepackage{tabularx}
\usepackage{subcaption}
\usepackage{bbding}
\usepackage{array}
\newcolumntype{C}[1]{>{\centering\arraybackslash}p{#1}}

\begin{document}

\begin{frontmatter}

\title{SUGFW+: An Uncertainty-guided Feature Weighting Framework for Cold Start Active Adaptation of SAM in Medical Image Segmentation}%

\author[1]{Xiaochuan Ma}
\author[2]{Ning Zhu}
\author[1]{Jia Fu}
\author[1,3]{Lanfeng Zhong}
\author[4]{Hanyu Jiang}
\author[4]{Bin Song}

\author[1]{Kang Li\corref{cores}}
\ead{kangli@uestc.edu.cn}

\author[1,3]{Guotai Wang\corref{cores}}
\ead{guotai.wang@uestc.edu.cn}

\cortext[cores]{Corresponding authors}

\address[1]{School of Mechanical and Electrical Engineering, University of Electronic Science and Technology of China, Chengdu, China.}
\address[2]{Glassgow College, University of Electronic Science and Technology of China, Chengdu, China.}
\address[3]{Shanghai Artificial Intelligence Laboratory, Shanghai, China.}
\address[4]{Department of Radiology, West China Hospital, Sichuan University, Chengdu, China.}

\begin{abstract}
Cold Start Active Learning (CSAL) is important in improving the performance of a medical image segmentation model with low annotation budget by querying a small subset for annotation from an unlabeled training set. Existing CSAL methods typically rely on inefficient dataset-specific Self-Supervised Learning (SSL) to map the unlabeled images into a feature space for sample selection. Recently, the advent of foundation models such as the Segment Anything Model (SAM) offer a promising alternative as the pre-trained model can provide strong generalizable feature embeddings, and allow high performance in downstream tasks after fine-tuning (adaptation). However, how to systematically exploit SAM's inherent embeddings for cold-start sample selection during adaptation with low annotation budget remains underexplored.  To address this, we propose an extended SAM-based Uncertainty-guided Feature Weighting (SUGFW+) framework for CSAL and adaptation of SAM. Specifically, it leverages the SAM for Patch-level Feature and Uncertainty Calculation (PFUC), and introduces a Patch-based Global Distinct Representation (PGDR) module that aggregates patch-level embeddings into highly discriminative, uncertainty-aware image-level features. These features are then utilized by a Greedy Selection with Cluster and Uncertainty (GSCU) strategy to combine diversity and uncertainty during sample selection. Unlike prior CSAL methods that decouple sample selection from model training, SUGFW+ tightly integrates these two stages via an Uncertainty-Prompted Fine-Tuning (UPFT) process of SAM in model training. Extensive experiments on four public datasets demonstrate that SUGFW+ achieves state-of-the-art performance against existing CSAL methods. Code is available at \href{https://github.com/HiLab-git/SUGFW-plus}{https://github.com/HiLab-git/SUGFW-plus}.

\end{abstract}

\begin{keyword}
Cold start active learning \sep
Domain adaptation \sep
Sample selection \sep
Uncertainty estimation
\end{keyword}

\end{frontmatter}

% \linenumbers

\section{Introduction}
Deep learning models have achieved strong performance in medical image segmentation with large amount of annotated data, but dense annotations required by medical image segmentation tasks are labor-intensive and time-consuming to obtain~\citep{mazurowski2023segment}. Active learning (AL)~\citep{AL_survey,luth2025nnactive} can reduce the annotation budget while maintaining competitive performance by strategically selecting only a small set of informative samples for annotation. Since different subsets may capture distinct aspects of the data distribution, the choice of annotated samples plays a critical role in ensuring the model performance. 
Existing AL methods focus primarily on two core properties: representativeness (diversity) and uncertainty. Representativeness-based methods~\citep{parvaneh2022active,bae2024generalized} aim to make the selected subset statistically consistent with the global data distribution, whereas uncertainty-based methods~\citep{thoma2025uncertainty,fuchsgruber2024uncertainty} identify samples for which the model yields ambiguous predictions. Additionally, hybrid approaches~\citep{bae2025uncertainty,he2024hybrid,luo2024uncertainty,zhong2025unisal} have been developed to synergistically integrate both criteria.
However, conventional AL methods suffer from a critical limitation at the initial stage: they typically require a small annotated set in the target dataset to initialize the model before meaningful selection. Furthermore, traditional AL relies on multi-round queries, which is computationally expensive and time-consuming for annotators. If constrained to a single query round to reduce both annotation and training time, these methods struggle to select an effective query batch for obtaining high model performance.

Cold Start Active Learning (CSAL)~\citep{zhu2025medcal} aims to address the cold start or single round sample selection problem without any exiting label the target dataset. While existing approaches commonly adopt Self-Supervised Learning (SSL)~\citep{chen2020simple,he2020momentum} coupled with traditional active learning selection strategies~\citep{Calr,Fps,Typiclust,Probcover,nguyen2022measure}, training an SSL model for each segmentation task is computationally expensive. To overcome this limitation, foundation models have recently been developed as robust pretrained feature extractors~\citep{wan2023survey,zhu2025medcal,Cec} that can be directly used for feature embedding of a downstream dataset, which avoids dataset-specific pretraining and therefore improves both feature generalizability and efficiency.
Compared with other  pretrained models such as DINO~\citep{dino2021},  the Segment Anything Model (SAM)~\citep{kirillov2023segment} is specifically designed for image segmentation tasks, and the corresponding feature representations have potential to improve sample selection effectiveness in CSAL of  segmentation models. In addition, the zero-shot inference capability of SAM allows obtaining a class-agnostic segmentation result, which provides more semantic information that can benefit sample selection for segmentation. 

%has emerged as a highly attractive solution due to its powerful zero-shot capabilities and strong feature representation ability.  As a result, SAM provides a potential for explicitly encoding pixel-level and spatial information, enabling more representative and diverse sample selection.

Beyond the potential for sample selection, foundation models have been shown to be strong pre-trained backbones, allowing achieving high  performance  in various downstream tasks after fine-tuning with a small set of labeled samples~\citep{ZHANG2024102996}. While Parameter-Efficient Fine-Tuning (PEFT) strategies like LoRA~\citep{hu2022lora} and adapters~\citep{chen2024ma} have been widely used for this purpose, %cheng2023sammed2d 
their efficacy is often constrained in low-budget regimes where scarce annotations trigger severe overfitting or representation degradation. Especially in CSAL setting, due to limited annotation budget, the number of labeled samples after querying is extremely low in the downstream task dataset, which increases challenges for effectively fine-tuning foundation models like SAM in such scenarios. % To mitigate these adverse effects, we propose integrating the inherent uncertainty of the selected samples to explicitly regularize the adaptation process. This uncertainty-prompted fine-tuning ensures a more robust and reliable transition from SAM’s generalized knowledge to domain-specific medical imaging expertise.

To address the above issues and improve CSAL performance under low annotation budget in medical image segmentation, we propose an extended SAM‑based Uncertainty‑guided Feature Weighting (SUGFW+) framework that leverages SAM for effective sample selection and model training in CSAL. %, which integrates two main modules: 1) SAM-guided cold start sample selection; 2) Uncertainty-prompted SAM fine-tuning. 
First, for the unannotated downstream dataset, SAM was used to obtain patch‑level image features alongside with  uncertainty estimation. Then, an image-level feature representation is obtained by  uncertainty-weighted integration of patch-level features. %To enhance the discriminative power of these features, we introduce the Patch‑based Global Distinct Representation (PGDR) module, which integrates patch‑wise features with uncertainty to produce a more representative feature space. 
Subsequently, querying samples are selected in the uncertainty-aware feature space by considering each sample's uncertainty and representativeness simultaneously. % hese representations are utilized by the most informative and diverse samples for manual annotation, thereby balancing uncertainty and representativeness. 
Finally, to leverage these selected samples to train a downstream segmentation model, we further perform an uncertainty‑prompted fine‑tuning strategy that adapts SAM to the downstream task by using uncertainty as a prompt to drive the model to be aware of hard regions for more effective learning. The  contribution of this work is three-fold:
\begin{itemize}
    \item We introduce SUGFW+, a novel CSAL framework that utilizes SAM’s feature representation with uncertainty estimation for effective sample selection and downstream fine-tuning, leadint to both high annotation efficiency and segmentation performance for a new unannotated dataset.
    \item To optimize sample selection, we introduce a SAM-based Patch-level Feature and Uncertainty Calculation (PFUC) method, and a Patch-based Global Distinct Representation (PGDR) strategy to obtain a uncertainty-aware global feature representation, followed by a Greedy Selection with Clustering and Uncertainty (GSCU) strategy that combines SAM-derived uncertainty and representativeness for selecting querying samples.
    \item To effectively train a downstream segmentation model with the query set, we propose an Uncertainty-Prompted Fine-Tuning (UPFT) method that adapts SAM to the downstream task with explicit uncertainty awareness for more informative learning. 
\end{itemize}
 
This work is a substantial extension of our preliminary conference publication~\citep{ma2025sugfw} where  PGDR  and GSCU were introduced to select querying samples for training a standard UNet backbone. This work significantly extended it in two key aspects: 
1) We replace the UNet backbone with SAM and design an uncertainty-prompted fine-tuning method UPFT, so that SAM's pre-trained representations are used for both sample selection and model training in a unified framework. 
%2) We replace Gini Impurity with Shannon entropy for uncertainty estimation, yielding a continuous, fine-grained uncertainty signal that better discriminates informative samples in high-uncertainty regions. 
2) We substantially expand the experimental evaluation to more medical image segmentation datasets, and comparisons against more recent state-of-the-art CSAL methods. 
Ultimately, extensive experiments across these four medical image segmentation datasets demonstrate that the proposed SUGFW+ achieves state-of-the-art performance, and even outperforms fully supervised UNet while reducing the annotation cost to 0.1\% to 3.0\% on these datasets.

\section{Related work}
\subsection{Cold start active learning}
Active learning (AL) aims to reduce annotation cost by iteratively selecting the most informative unlabeled samples for labeling. Early AL methods are broadly categorized into three streams. Uncertainty-based methods select samples for which the current model is least confident, using metrics such as entropy, margin sampling, or Bayesian disagreement~\citep{gal2016dropout,beluch2018power}. Diversity-based methods aim to cover the feature space by selecting a representative subset, often via clustering or core-set selection~\citep{sener2018active,sinha2019variational}. Hybrid approaches combine both criteria, balancing exploration of the feature space with exploitation of model uncertainty~\citep{ash2020deep}. However, these conventional AL methods typically require an initial annotated set to warm up the model, and their reliance on multi-round queries incurs prohibitive time and computational costs.

To address these limitations, Cold Start Active Learning (CSAL) aims at selecting the most useful subset of training images for annotation from an unlabeled dataset. Unlike general AL that has an initialized model trained with few labeled samples to guide sample selection, CSAL is more challenging due to the unavailability of such an initial model. Traditionally, many methods rely on SSL to extract image features and then select samples with diversity or uncertainty~\citep{Fps,Calr,Typiclust}. % ,Hacon,Probcover
For example, \cite{Fps} utilized a Farthest Point Selection (FPS) strategy to ensure uniform distribution of selected samples. \cite{Calr} employed birch hierarchical clustering and selected the samples with the highest information density from each cluster. \cite{Typiclust} selected the sample with the highest density within each cluster, ensuring that the chosen points are less affected by clustering noise. \cite{Probcover} selected the most representative initial samples by iteratively choosing points that cover the largest number of uncovered samples within a fixed radius. % $\sigma$. 
\cite{Hacon} selected the hardest-to-contrast samples within each cluster after SSL as typical data. %the most representative and class-diverse initial samples by choosing , which utilized the SSL model in sample selection. 
Recently, some researchers have turned to foundation models for feature extraction. \citep{zhu2025medcal,Cec,ALPS} in CSAL. Most of them simply use the pre-trained feature extractors  to obtain representations of unlabeled samples, and use traditional clustering and representativeness-based methods for sample selection. However, the ignorance of uncertainty information in these methods could lead to limited performance.  In contrast, our method exploits the zero-shot inference ability of SAM to produce region-level meaningful uncertainty estimations before any target-domain training for more informed query. %In addition, our method further unifies sample selection and model adaptation into an uncertainty-driven pipeline, where the uncertainty that guides selection also directs the subsequent fine-tuning.

\subsection{Adaptation of SAM for medical image segmentation}
The adaptation of the SAM~\citep{kirillov2023segment} to medical image segmentation has been extensively investigated in recent years~\citep{ali2025review}. Early studies primarily focused on preserving SAM’s strong prompt-driven segmentation paradigm. For instance, \cite{dai2023samaug} proposed an augmented point-prompt strategy to improve segmentation performance, while \cite{li2024assessing} demonstrated that selecting a point near the center of a lesion or organ yields superior segmentation results. Considering the gap between pre-trained features and downstream segmentation tasks, several works aims to adapt SAM for medical image segmentation through task-specific fine-tuning. SAM-Med2D~\citep{cheng2023sammed2d} freezes the image encoder while introducing trainable adapter layers, refining the prompt encoder, and modifying the mask decoder during training. Similarly, MedSAM~\citep{MedSAM} tailors SAM for medical image segmentation by keeping the prompt encoder fixed and fine-tuning both the image encoder and the mask decoder. Despite their effectiveness, these methods still rely on human-provided prompts, such as points or bounding boxes, which limit their scalability and practical applicability. To move toward fully automatic segmentation, MA-SAM~\citep{chen2024ma} inserts adapters and LoRA layers into the image encoder’s transformer blocks to capture medical imaging contexts and fully fine-tunes the mask decoder. 
Nonetheless, the existing SAM adaptation methods operate in a fully‑supervised setting, which requires a relatively large set of annotated images with high annotation cost. In contrast, this work deals with  SAM adaptation under a CSAL scenario, where only a very small initial batch of images are suggested for annotation, and we leverage the  uncertainty information derived from SAM's features and predictions for both informative sample selection and efficient model adaptation,  enabling the model to be effectively optimized with a minimal annotation budget.

\begin{figure*}[t!]
  \centering
  \centerline{\includegraphics[width=18cm]{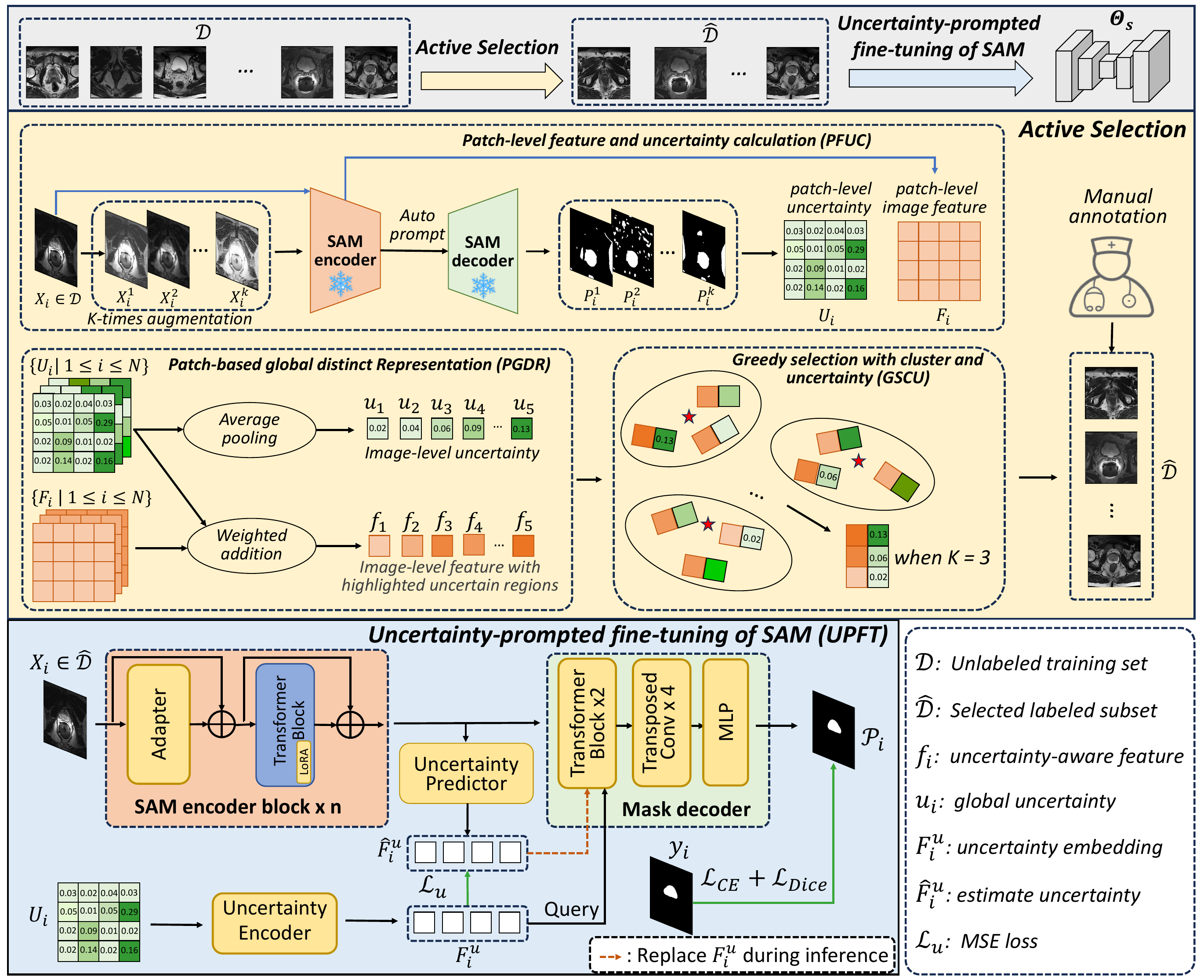}}
\caption{
Overview of our proposed SUGFW+ framework for CSAL in medical image segmentation. It consists of two core components: 1) SAM-based Uncertainty-Guided Feature Weighting (SUGFW) for selecting a small set of query samples from the  unannotated target dataset, and 2) Uncertainty-Prompted Fine-Tuning (UPFT) of SAM with the query set for model training.  SUGFW consists of three steps: 1) SAM-based Patch-level Feature and Uncertainty Calculation (PFUC), 2) Patch-based Global Distinct Representation (PGDR) and 3) Greedy Selection with Cluster and Uncertainty (GSCU). %SUGFW diversity and uncertainty-aware sample selection to identify informative annotations, and uncertainty-prompted adaptation to fine-tune SAM with limited labels. Both capabilities are built upon SAM-derived features and uncertainty. To achieve sample selection, the ``everything mode'' of SAM is deployed on multiple augmented images to generate predictions, followed by uncertainty estimation and feature extraction. Then we introduce the PGDR strategy to derive distinct feature representations and quantify uncertainty, coupled with the GSCU strategy to select samples that are both representative and exhibit uniformly distributed uncertainty for manual annotation. For uncertainty-prompted fine-tuning, we incorporate adapters and LoRA modules into SAM's image encoder, while explicitly integrating uncertainty to prioritize reliable regions and enhance training robustness. Additionally, an uncertainty predictor is trained to estimate uncertainty embeddings to eliminate the computational overhead of uncertainty extraction during inference.
} 
\label{fig::structure}
\end{figure*}

\section{Method}
\subsection{Method overview}
Considering an unlabeled training set $\mathcal{D} = \{X_i\}_{i=1}^{N}$ with $N$ samples, we aim to select a subset $\mathcal{\hat{D}}=\{X_{j}\}_{j=1}^M$ with $M$ samples for annotation in a single query that are then used for model training, where $M\ll N$. The challenge is twofold: first, the selected $\mathcal{\hat{D}}$ should be highly representative of $\mathcal{D}$ without any label guidance; second, the subsequent fine-tuning on $\mathcal{\hat{D}}$ should be effective enough to achieve strong performance despite the limited annotated samples. 

The overall structure of our framework SUGFW+ for CSAL is illustrated in Fig.~\ref{fig::structure}. It consists of two main steps: 1) SAM-based Uncertainty-Guided Feature Weighting (SUGFW) for cold-start sample selection (i.e., querying for annotation), and 2) Uncertainty-Prompted Fine-Tuning (UPFT) of SAM with the small set of query samples. % Moreover, the chosen subset is employed to train an automatic segmentation model by adapting SAM with an UPFT strategy to obtain high segmentation performance.

\subsection{SUGFW for sample querying}
In SUGFW, a Patch-level Feature and Uncertainty Calculation (PFUC) module is first used to obtain patch-level image features and uncertainty by applying the ``everything mode'' of SAM~\citep{kirillov2023segment} with $K$-time augmentation. Subsequently, we employ a Patch-based Global Distinct Representation (PGDR) strategy to derive distinctive image-level representations. Then, a Greedy Selection with Cluster and Uncertainty (GSCU) strategy is employed to select samples that are both representative and informative for annotation.
\subsubsection{Patch-level feature and uncertainty calculation}
CSAL requires a feature extractor to map the unannotated dataset into a feature space. Compared with pretrained encoders like those in CLIP and DINO that obtain high-level task-agnostic feature representations, SAM~\citep{kirillov2023segment} is more appealing  due to not only that the encoder is pre-trained with segmentation tasks, but also that its decoder can obtain segmentation results to provide more semantic information and uncertainty estimation that can assist sample selection in CSAL.  %,  the  method to characterize the unlabeled dataset, yet existing SSL-pre-trained encoders are trained with instance-level objectives and do not naturally provide per-pixel uncertainty signals. SAM~\citep{kirillov2023segment} addresses both needs: beyond strong image features from large-scale pretraining, its class-agnostic mask outputs directly reflect local prediction confidence and can serve as a natural uncertainty source. 
Therefore, we propose Patch-level Feature and Uncertainty Calculation (PFUC) based on SAM,  which leverages the SAM's image encoder $E_{img}$ to obtain patch-level features, and take advantage of the masks predicted by a mask decoder  $D_{m}$ conditioned on a prompt encoder $E_{p}$ to obtain uncertainty estimation of each patch. %which consists of an image encoder $E_{img}$ and a mask decoder $D_{m}$ for predicting segmentation masks. Additionally, a prompt encoder $E_{prom}$ is incorporated to align image and prompt features.

Specifically, for a given image $X_{i} \in \mathcal{R}^{H\times W}$, where $H$ and $W$ denote the image size,  the feature map corresponding to the output of $E_{img}$ is denoted as:
\begin{equation}
F_i=E_{img}(X_{i}),
\label{eq:feature_map}
\end{equation}
where $F_{i} \in \mathcal{R}^{C_0\times H'\times W'}$, with $C_0$ being the feature dimension and $H'\times W'$ being the feature map size, respectively. 
Given a  prompt $\mathcal{P}$, the preliminarily predicted mask $M'_{i}$ for $X_i$ is:
\begin{equation}
M'_{i} = D_{m}\big(E_{img}(X_{i}), E_{p}(\mathcal{P})\big),
\label{eq1}
\end{equation}

To avoid human intervention during the sample selection stage, % and   assess SAM's comprehension of an image without annotations, 
we utilize SAM's ``everything mode'', where SAM generates a grid of points as   the prompt $\mathcal{P}$ to obtain a segmentation output that is a set of class-agnostic regions. 
To reduce noise and reject background regions, we take a union of the segmented regions with a size smaller than half of the image, obtaining a binary foreground segmentation mask $M_{i}$, which is formulated as:
\begin{equation}
M_{i} = \bigcup_{n=1}^{N_i} M'_{i}(n)\mathcal{I}[|M'_{i}(n)| < HW/2] ,
\label{eq2}
\end{equation}
where $N_i$ is the number of regions in $M'_{i}$, and $M'_i(n)$ represents the size of the $n$-th region. 
$\mathcal{I(\cdot)}$ is the indicator function that returns 1 if the condition is satisfied and 0 otherwise, $|M'_i(n)|$ represents the size of $M'_i(n)$. %$H$ and $W$ denote the height and width of $X_{i}$, respectively. 

Furthermore, to obtain an  uncertainty estimation of  the image, we first apply $K$ augmentations to $X_{i}$, including intensity and spatial transformations. Subsequently, we utilize Eq.~\eqref{eq1} and Eq.~\eqref{eq2} to derive the segmentation results $M_i^k$ for the $k$-th augmentation. By taking an average of the $K$ predictions, a soft prediction mask is denoted as:
\begin{equation}
\bar{M}_{i} = \frac{1}{K} \sum_{k=1}^{K} M_{i}^{k}
\end{equation}

As the size of $\bar{M}_{i}$ is $H\times W$, we down-sample it to $H' \times W'$ to match the size of feature map $F_i$, and denote the down-sampeled version of $\bar{M}_{i}$ as $\tilde{M}_{i}$.  
From $\tilde{M}_{i}$, the softness of the mask value for each element can represent the uncertainty of the corresponding image patch, leading to an uncertainty map $U_{i} \in \mathcal{R}^{H' \times W'}$, and the $r$-th element of $U_{i}$ is obtained by calculating the entropy, which is formulated as:
\begin{equation}
U_{i}(r) = -[\bar{M}_{i}(r)log\bar{M}_{i}(r) + \big(1-\bar{M}_i(r)\big)log\big(1-\bar{M}_i(r)\big)],
\label{eq3}
\end{equation}
where %$r$ is the  $\bar{M}_{i} = \frac{1}{K} \sum_{k=1}^{K} M_{i}^{k}$ to average all masks. 
a lower $U_i(r)$ indicates that SAM has a more reliable understanding of the corresponding  region under the perturbations added to the input,  corresponding to a lower uncertainty. Note that this SAM-derived uncertainty cannot totally represent the uncertainty relevant to the target segmentation task, so we take $U_i$ as a surrogate uncertainty map.  

%Note that the uncertainty derived from SAM’s ``everything mode'' reflects its intrinsic representative capacity across various image patches. This fundamentally differs from conventional predictive uncertainty, which is typically estimated by models trained on specific supervised tasks.
 
\subsubsection{Patch-based global distinct representation}
According to the feature map $F_i$ and surrogate uncertainty map $U_i$, the patch-level features in $F_i$ should be aggregated into a global image-level feature representation for the following sample selection step. Therefore,  we propose a Patch-based Global Distinct Representation (PGDR) strategy, where the patch-level image features are weighted and fused based on their  surrogate uncertainty, resulting in an uncertainty-aware global  feature $f_{i}$: % where the uncertain regions are amplified:
\begin{equation}
f_{i} = \sum_{r=1}^{H'\times W'}  F_{i}(r) \cdot \frac{e^{\lambda\cdot U_{i}(r)}}{\sum_{r=1}^{R}e^{\lambda \cdot U_{i}(r)}},
\label{eq:global_feature}
\end{equation}
where $\lambda \geq 1.0$ is a hyper-parameter that amplifies the contribution of high-uncertainty regions, emphasizing informative samples for the subsequent selection stage. 

Alongside $f_i$, we also aggregate the patch-level uncertainty in $U_i$ into an image-level surrogate uncertainty score $u_i$ by applying average pooling to $U_i$:
\begin{equation}
u_{i} = \frac{1}{H'\times W'}\sum_{r=1}^{H'\times W'} U_{i}(r),
\label{eq5}
\end{equation}
%. Then, Global average pooling is applied to $U_{i}$ to obtain $\hat{U}_{i}$, which represents the image-level uncertainty. $\hat{F}_{i}$ and $\hat{U}_{i}$ are obtained by:

% Through the PGDR strategy, we obtain distinct representations $\hat{F}_{i}$ and $\hat{U}_i$ for $X_i$, which are then used in sample selection.

% Through the PGDR strategy, we ultimately obtain more distinctive global representations $\hat{F}_{i}$ and $\hat{U}_{i}$. 

\subsubsection{Greedy selection with cluster and uncertainty}
With the image-level global representation $f_{i}$ and the  surrogate uncertainty $u_{i}$, to ensure that the selected samples are both informative and representative, we propose a Greedy Selection with Cluster and Uncertainty (GSCU) strategy. Firstly, the K-means clustering algorithm is applied to the image-level features $f_{i}$, and the cluster number $C$ is the same as the number of samples to be selected, i.e., annotation budget in CSAL. For $X_{i}$, its assigned class $c_{i}$ is obtained by:
\begin{equation}
c_{i} = \arg\min_{c \in \{1, 2, \dots, C\}} \| f_{i} - \mu_c \|^2,
\label{eq6}
\end{equation}
where $C$ represents the number of clusters and $\mu_c$ denotes the centroid of the cluster. Then, we select one sample from each cluster to get the query batch. Here, the strategy to select the sample in each cluster plays an important role for the effectiveness of querying. First, simply selecting the centroid of each cluster will focus too much on the representativeness while ignoring uncertainty of $f_i$. Second, selecting the most uncertain sample from each cluster may also be suboptimal as the $u_i$ value is just an estimation rather than accurate measurement of the uncertainty for the target segmentation task, due to the class-agnostic segmentation output of SAM. To deal with this problem, our GSCU strategy  encourages diversity of surrogate uncertainty while keeping feature representativeness for sample selection. %takes uncertainty into consideration so that  we adopt greedy selection (GS) that enforces diversity in the uncertainty dimension.
For the first querying sample, we select the one with the median surrogate uncertainty in a random cluster. For subsequent selections, a sample is greedily selected from another random cluster so that its minimum distance to the surrogate uncertainties of all currently selected samples is maximized. Specifically, let $\mathcal{D}^{m}$ denote the set of query samples after the $m$-th selection, %a subset $\mathcal{D}'$ is extracted from the training set $\mathcal{D}$. When 
when selecting one more query  sample from a new cluster $\mathcal{C}$, the ($m+1$)-th query sample is determined by:% . A sample $X_{i}'$ from $\mathcal{C}$ will be chosen according to the following criteria:
\begin{equation}
X_{m+1} = \arg\max_{X_i \in \mathcal{C}} (\min_{X_j \in \mathcal{D}^m} \left| u_i - u_j \right| ),
\label{eq7}
\end{equation}

Through this GSCU strategy, we obtain a subset $\hat{\mathcal{D}}$ with $M$ querying samples that represents the distribution of the training set in both feature space and the surrogate uncertainty space. The chosen samples are then annotated by the human annotator. After that,  we use supervised training on these samples in $\hat{\mathcal{D}}$ to train the task-specific automatic segmentation model, as detailed in the following. 

\subsection{Uncertainty-prompted  fine-tuning of SAM}
Rather than training a separate segmentation model from scratch, we adapt SAM itself as the downstream model to enhance the performance by leveraging its pre-trained weights. %In addition, this ensures that the features used for sample selection and those used for segmentation reside in the same representation space, allowing the uncertainty estimated during selection to directly guide model adaptation. 
Existing works on PEFT based on adapters~\citep{chen2024ma} and LoRA~\citep{hu2022lora} have shown high performance for fine-tuning foundation models. However, in this work, there are two unique considerations that are different from standard PEFT or SAM adaptation settings. First, in our CSAL setting, the size of $\hat{\mathcal{D}}$ is much smaller than that of annotated image set for common PEFT, making the fine-tuning more challenging. Second, SAM is originally an interactive segmentation model, while we aim to obtain an automatic segmentation model after the PEFT. To deal with these problems, we propose Uncertainty-Prompted Fine-Tuning (UPFT) of SAM in downstream model training. 

% into each transformer layer of the image encoder, which preserves SAM's pre-trained knowledge while enabling task-specific adaptation with minimal additional parameters. During early training, the model's predictions are unstable, making it difficult to identify which regions require more attention. 
Specifically, we use the surrogate uncertainty map $U_i$ estimated from SAM's mask outputs as a spatial prompt, which avoids manual prompting and provides informative guidance for model adaptation. %directing the model to focus on high-uncertainty regions where its predictions are least confident. This uncertainty-guided signal is integrated into the fine-tuning process to accelerate adaptation on the target dataset.
The SAM's original prompt encoder is therefore replaced by an uncertainty encoder  $E_{un}$ to obtain an uncertainty embedding $F_i^u = E_{un}(U_i)$. Then $F_i^u$ is injected into the mask decoder $D_m$, and used as quires in the cross attention blocks in the decoder as implemented in the original SAM architecture. 
%Specifically, we first invert the uncertainty map to obtain $U_i^*$, where $U_i^*=1-\hat{U_i}$. Then we employ an uncertainty encoder $E_{un}$ to align the dimensions of the uncertainty embeddings with the decoder's input requirements, which is denoted as $F_i^u = E_{un}(U_i^*)$. These aligned embeddings are subsequently used as queries during the decoding stage to increase the attention weights assigned to reliable regions. 
% This design encourages the model to focus more on reliable areas during training, thereby facilitating more effective learning. 
Note that calculating $U_i$ requires $K$ forward passes of SAM, which is time-consuming and computationally expensive for inference. To deal with this problem, we further introduce a lightweight uncertainty predictor   $g(\cdot)$ that directly estimates the uncertainty embedding based on $F_i$:
\begin{equation}
    \hat{F_i^u} = g(F_i, \bar{U})),
\label{eq:uncertainty_pred}
\end{equation}
where $\bar{U}$ is the average surrogate uncertainty map across the training set $\mathcal{D}$, and $g(\cdot)$ is implemented as a lightweight transformer and trained with the Mean Squared Error (MSE) loss:
\begin{equation}
    \mathcal{L}_u=||\hat{F_i^u} - F^u_i||^2_2,
\end{equation}

During the training stage, the segmentation model's prediction for $X_i \in \hat{\mathcal{D}}$ is prompted by $F^u_i$, and denoted as:
\begin{equation}
    {P}_i = D_{m}\big(E_{img}(X_i), F_i^u\big),
\label{eq:seg_predict}
\end{equation}
where ${P}_i$ is supervised by the manual annotation through a Dice loss $\mathcal{L}_{Dice}$ and a cross entropy loss $\mathcal{L}_{CE}$ as used in standard supervised training. The total loss for model training is: 
%Let $E_{m}$ denote the SAM's encoder augmented with adapter and LoRA injection mechanisms, $D_{seg}$ represent the mask decoder, and $E_{un}$ signify the uncertainty encoder. For an input image $X_i$ with the annotation $y_i$, the predicted mask $\mathcal{P}_i$ is formulated as follows:

%The Dice and cross-entropy loss are utilized as the optimization objective, which are formulated as: 
%\begin{equation}
%    \mathcal{L}_{Dice} = \frac{2\sum_{}^{}\mathcal{P}_i\cdot y_i + \epsilon}{\sum \mathcal{P}_i^2 + \sum y_i^2 + \epsilon},
%\end{equation}

%\begin{equation}
%    \mathcal{L}_{CE} = \sum [y_i \cdot log(\mathcal{P}_i) + (1 - \mathcal{P}_i) \cdot y_i],
%\end{equation}
%where $\epsilon$ is a bias to avoid division by zero.

%Computing $U_i^*$ requires $K$ forward passes through SAM, which is impractical during inference. To eliminate this overhead, we learn a lightweight uncertainty predictor $g(\cdot)$ that directly estimates the uncertainty embedding from a single forward pass. During training, $g(\cdot)$ is optimized to estimate $\hat{F_i^u}$:
% Considering the initial instability of $F_i^u$, we employ a warm-up strategy that gradually ramps up the learning rate of the MSE loss.
%The total loss $\mathcal{L}$ is formulated as:
\begin{equation}
    \mathcal{L}=\omega \cdot \mathcal{L}_{Dice} + (1 - \omega) \cdot \mathcal{L}_{CE} + \alpha_t\cdot \mathcal{L}_{u}.
    \label{eq:total_loss}
\end{equation}
where $\omega$ controls the balance between $\mathcal{L}_{Dice}$ and $\mathcal{L}_{CE}$. $\alpha_t = min(\alpha_{max},\alpha_{max} \cdot \frac{t}{T})$ is the weight for $\mathcal{L}_u$ set by a ramp-up strategy, with $\alpha_{max}$  being the maximum weight and $T$ being the the warm-up iterations. During model adaptation, we add LoRA layers to the image encoder $E_{img}$, and $\mathcal{L}$ is backpropagated to update the LoRA layers, mask decoder $D_m$, uncertainty encoder $E_{un}$ and the uncertainty predictor $g(\cdot)$. 
 During inference, $E_{un}$ is discarded, and we use Eq.~\eqref{eq:uncertainty_pred} to obtain the surrogate uncertainty embedding $\hat{F}^u_i$, and it replaces $F^u_i$ in Eq.~\eqref{eq:seg_predict} to obtain the segmentation output.  
 %$\hat{F}_i^u$ directly replaces $F_i^u$, removing the need for repeated SAM forward passes.

\begin{table*}[t!]
  \centering
  \caption{Quantitative comparison of different CSAL methods on the Promise12 dataset. The best results are in bold, and the second-best are underlined. $*$ indicates that the p-value $<$ 0.05 in paired t-test compared to the second-best results.} 
  \resizebox{\textwidth}{!}{
    \begin{tabular}{p{1cm}|c|cccccccc}
    \toprule
    \multirow{2}{*}{Metric} & \multirow{2}{*}{Method} & \multicolumn{6}{c}{Annotation Budget} \\
    % \cmidrule(lr){3-8}
    & & 3.0\%    & 4.0\%    & 5.0\%    & 6.0\%    & 7.0\%    & 8.0\% \\
    \midrule
    \multirow{8}{*}{\makecell{DSC \\ (\%) $\uparrow$}} & Random & 84.03±5.87 & 85.72±4.86 & 85.74±4.33 & 86.24±6.79 & 86.55±6.27 & 86.64±6.60 \\
          & ALPS~\citep{ALPS} & 81.43±8.39 & 82.46±10.14 & 84.97±7.09 & 86.44±5.37 & 87.92±4.48 & 88.39±5.65 \\
          & CALR~\citep{Calr} & 82.89±5.74 & 84.44±6.90 & 85.72±5.59 & 86.63±5.56 & 86.84±7.80 & 87.53±7.08 \\
          & FPS~\citep{Fps} & \underline{84.24±6.15} & 85.40±5.48 & \underline{86.95±6.31} & \underline{87.16±4.48} & 87.32±5.50 & 88.21±3.56 \\
          & Probcover~\citep{Probcover} & 83.08±4.82 & \underline{86.22±4.82} & 86.49±5.26 & 86.70±4.15 & \underline{88.53±3.56} & \underline{88.88±3.52} \\
          & Typiclust~\citep{Typiclust} & 82.78±6.79 & 84.24±4.95 & 85.39±5.75 & 86.11±4.24 & 88.16±5.43 & 88.64±4.33 \\
          & CEC~\citep{Cec} & 82.41±7.69 & 83.62±8.30 & 85.40±6.17 & 85.60±4.21 & 86.14±4.88 & 86.53±5.32 \\
          & SUGFW+ (Ours) & \textbf{86.38±5.60}$^*$ & \textbf{87.35±3.90}$^*$ & \textbf{87.86±5.86}$^*$ & \textbf{88.51±3.92}$^*$ & \textbf{89.20±3.11}$^*$ & \textbf{89.25±3.10}$^*$ \\
    \midrule
    \multirow{8}{*}{\makecell{HD95 \\ ($mm$) $\downarrow$}} & Random & 3.74±3.12 & 3.67±7.62 & 3.24±3.22 & 2.41±4.51 & 1.98±1.91 & 1.79±1.37 \\
          & ALPS~\citep{ALPS} & 5.91±7.23 & 2.95±2.88 & 2.76±2.66 & \underline{2.00±1.74} & \underline{1.58±0.83} & \underline{1.56±2.22} \\
          & CALR~\citep{Calr} & 6.50±6.50 & 5.73±6.98 & 4.77±11.09 & 4.77±7.24 & 3.90±4.21 & 2.69±6.92 \\
          & FPS~\citep{Fps} & \underline{3.47±4.84} & 3.00±3.50 & \underline{2.02±1.59} & 3.00±4.80 & 1.73±1.54 & 1.83±1.43 \\
          & Probcover~\citep{Probcover} & 37.80±112.88 & \underline{2.12±1.46} & 3.00±3.73 & 3.74±4.44 & 1.71±1.35 & 2.97±5.39 \\
          & Typiclust~\citep{Typiclust} & 18.20±75.85 & 4.10±3.46 & 4.16±4.77 & 2.49±6.08 & 2.09±2.19 & 1.62±1.73 \\
          & CEC~\citep{Cec} & 13.70±24.87 & 6.19±8.04 & 4.98±4.61 & 3.05±3.61 & 2.71±2.17 & 2.34±2.11 \\
          & SUGFW+ (Ours) & \textbf{2.92±4.17}$^*$ & \textbf{2.10±1.91} & \textbf{1.94±1.73} & \textbf{1.58±0.95}$^*$ & \textbf{1.37±0.66}$^*$ & \textbf{1.35±0.67}$^*$ \\
    \bottomrule
    
    \end{tabular}
    }%
  \label{tab:main:Promise12}%
\end{table*}%

\section{Experiments and results}
\subsection{Datasets}
We evaluated our method in four public medical imaging datasets with various modalities, anatomy structures, and imaging protocols to assess its effectiveness and generalization. The datasets include:

  1) The \textbf{Promise12 dataset}~\citep{Promise12dataset} that is for prostate segmentation in T2-weighted Magnetic Resonance Imaging (MRI). It comprises 100 transverse MRI scans acquired from 6 clinical centers, with voxel spacing ranging from 0.4--0.75\,mm in-plane and 3--4\,mm slice thickness. Following the official split, the dataset is divided into training, validation, and test sets in a ratio of 5:2:3.

  2) The \textbf{UTAH dataset}~\citep{UTAHdataset} that consists of 154 3D MRI scans of atrial fibrillation patients %, predominantly provided by the University of Utah's NIH/NIGMS Center for Integrative Biomedical Computing with   additional contributions  (all with ethical approval)
  from multiple institutions. The images were scanned with an isotropic resolution of 0.625\,mm$^3$ , each accompanied by expert-annotated left atrial cavity segmentations.
  Following the official partition, the 3D images were divided into 80, 20 and 54 cases for training, validation and  testing, respectively. 

  3) The \textbf{MSD Liver} dataset~\citep{MSDLiverdataset} from the Medical Segmentation Decathlon (MSD) %is a widely recognized benchmark 
  for liver segmentation in contrast-enhanced abdominal Computed
  Tomography (CT). It comprises 131 3D CT volumes with highly variable spatial resolution and slice thickness, reflecting the heterogeneity of clinical imaging protocols in different institutions. Each volume is annotated with
  per-voxel segmentation masks of the liver parenchyma. We aim to segment the whole liver, %only, excluding intrahepatic tumors from the target. The 
and split the dataset into training, validation, and testing sets at a ratio of 7:1:2.

  4) The \textbf{ISIC 2018} dataset~\citep{ISICdataset} from the International Skin Imaging Collaboration for skin lesion segmentation.  It consists of 3,694 high-resolution 2D dermoscopic images   encompassing a wide spectrum of skin conditions, including melanomas, nevi, and seborrheic keratoses. %Each image is accompanied by histopathology-confirmed lesion diagnoses and expert-annotated segmentation masks.
  Following the
  official split, the dataset is divided into 2,595 images for training, 100 for validation, and 1,000 for testing.

\begin{table*}[t!]
  \centering
  \caption{Quantitative comparison of different CSAL methods on the UTAH dataset. The best results are in bold, and the second-best are underlined. $*$ indicates that the p-value $<$ 0.05 in paired t-test compared to the second-best results.}
  \resizebox{\textwidth}{!}{
    \begin{tabular}{p{1cm}|c|cccccccc}

    \toprule
    \multirow{2}{*}{Metric} & \multirow{2}{*}{Method} & \multicolumn{6}{c}{Annotation Budget} \\
    & & 0.10\%    & 0.15\%    & 0.20\%    & 0.25\%    & 0.50\%    & 0.75\% \\
    \midrule
    \multirow{8}{*}{\makecell{DSC \\ (\%) $\uparrow$}} & Random & 70.96±11.96 & 75.17±5.32 & \underline{81.51±4.84} & 82.51±4.06 & 85.00±3.58 & 87.67±3.23 \\
          & ALPS~\citep{ALPS} & 73.80±6.48 & 77.24±5.69 & 80.72±5.19 & \underline{83.73±4.58} & \underline{85.68±5.08} & 87.61±4.21 \\
          & CALR~\citep{Calr} & 73.78±5.59 & 74.96±5.40 & 81.28±4.47 & 80.72±4.89 & 82.84±5.72 & \underline{88.66±3.44} \\
          & FPS~\citep{Fps} & 46.97±12.54 & 69.74±13.30 & 70.16±14.79 & 71.58±10.94 & 71.76±7.62 & 82.57±7.74 \\
          & Probcover~\citep{Probcover} & 62.72±12.83 & 74.78±5.21 & 76.83±5.17 & 76.90±4.78 & 78.41±4.62 & 83.30±4.13 \\
          & Typiclust~\citep{Typiclust} & \underline{74.03±5.25} & 70.14±8.75 & 81.23±5.53 & 82.57±4.90 & 84.27±4.88 & 88.40±3.46 \\
          & CEC~\citep{Cec} & 73.40±9.38 & \underline{78.30±5.48} & 80.94±4.88 & 83.37±4.50 & 84.43±4.39 & 88.17±3.84 \\
          & SUGFW+ (Ours) & \textbf{78.70±5.66}$^*$ & \textbf{82.26±5.05}$^*$ & \textbf{84.02±4.99}$^*$ & \textbf{85.17±4.32}$^*$ & \textbf{86.94±4.02}$^*$ & \textbf{89.44±3.59}$^*$ \\
    \midrule
    \multirow{8}{*}{\makecell{HD95 \\ ($mm$) $\downarrow$}} & Random & 43.51±23.62 & 25.61±13.96 & 25.32±8.43 & 25.54±7.60 & 28.00±10.51 & 32.95±14.63 \\
          & ALPS~\citep{ALPS} & \underline{28.43±9.86} & 27.15±12.82 & 31.45±10.09 & \underline{17.82±7.55} & \underline{13.09±5.94} & 13.51±9.32 \\
          & CALR~\citep{Calr} & 46.42±38.06 & 26.94±20.19 & 26.17±8.86 & 24.92±8.13 & 20.02±5.80 & \underline{9.92±5.71} \\
          & FPS~\citep{Fps} & 43.85±12.66 & 30.40±10.96 & 29.05±7.67 & 28.35±10.14 & 27.35±7.88 & 15.27±5.29 \\
          & Probcover~\citep{Probcover} & 44.04±19.97 & \underline{21.21±5.72} & \underline{23.80±8.47} & 24.24±7.74 & 23.88±7.08 & 16.53±3.35 \\
          & Typiclust~\citep{Typiclust} & 28.86±11.31 & 30.22±14.86 & 24.86±8.16 & 20.26±10.24 & 26.39±7.93 & 10.87±7.23 \\
          & CEC~\citep{Cec} & 29.50±25.09 & 21.97±7.74 & 26.15±9.15 & 31.24±13.49 & 27.83±9.49 & 11.80±6.07 \\
          & SUGFW+ (Ours) & \textbf{24.86±19.11}$^*$ & \textbf{19.98±7.72}$^*$ & \textbf{17.03±3.74}$^*$ & \textbf{15.19±3.77}$^*$ & \textbf{12.39±9.63}$^*$ & \textbf{7.30±4.29}$^*$ \\
    \bottomrule

    \end{tabular}}%
  \label{tab:main:UTAH}%
\end{table*}%

\begin{table*}[t!]
  \centering
  \caption{Quantitative comparison of different CSAL methods on the MSD Liver dataset. The best results are in bold, and the second-best are underlined. $*$ indicates that the p-value $<$ 0.05 in paired t-test compared to the second-best results.}
  \resizebox{\textwidth}{!}{
    \begin{tabular}{p{1cm}|c|cccccccc}

    \toprule
    \multirow{2}{*}{Metric} & \multirow{2}{*}{Method} & \multicolumn{6}{c}{Annotation Budget} \\
     &  & \multicolumn{1}{c}{0.05\%} & \multicolumn{1}{c}{0.06\%} & \multicolumn{1}{c}{0.08\%} & \multicolumn{1}{c}{0.10\%} & \multicolumn{1}{c}{0.20\%} & \multicolumn{1}{c}{0.30\%} \\
    \midrule
    \multirow{8}[2]{*}{\makecell{DSC \\ (\%) $\uparrow$}} & Random & 68.99±9.51 & 70.09±6.85 & 74.32±6.12 & 77.86±5.54 & 84.34±4.99 & 90.37±4.71 \\
          & ALPS~\citep{ALPS} & 63.79±9.30 & \underline{78.90±4.70} & \underline{83.17±4.28} & \underline{85.64±4.02} & 90.99±3.80 & \underline{92.57±3.31} \\
          & CALR~\citep{Calr} & 62.19±11.36 & 67.35±9.91 & 71.24±8.43 & 75.49±7.16 & 84.24±5.00 & 91.97±5.29 \\
          & FPS~\citep{Fps} & 72.15±5.07 & 75.58±6.01 & 76.13±5.89 & 76.82±5.67 & 77.98±5.45 & 89.58±4.06 \\
          & Probcover~\citep{Probcover} & 58.70±9.22 & 77.13±4.18 & 78.64±5.12 & 80.37±6.33 & 83.03±7.66 & 84.15±6.23 \\
          & Typiclust~\citep{Typiclust} & \underline{77.07±6.72} & 77.16±6.83 & 81.49±6.21 & 84.93±5.52 & \textbf{91.58±4.78} & 92.34±4.30 \\
          & CEC~\citep{Cec} & 52.58±10.17 & 58.25±10.03 & 61.47±9.14 & 63.58±8.22 & 68.52±6.58 & 90.70±4.96 \\
          & SUGFW+ (Ours) & \textbf{82.27±4.67}$^*$ & \textbf{87.94±3.61}$^*$ & \textbf{89.26±4.17}$^*$ & \textbf{90.41±4.26}$^*$ & \underline{91.53±4.51} & \textbf{93.11±4.65}$^*$ \\
    \midrule
    \multirow{8}[2]{*}{\makecell{HD95 \\ (mm) $\downarrow$}} & Random & 31.76±11.87 & 24.57±10.72 & 21.50±10.28 & \underline{18.61±10.12} & 15.01±9.61 & 12.18±7.87 \\
          & ALPS~\citep{ALPS} & 27.18±9.35 & 25.65±10.65 & 23.04±11.08 & 21.13±11.45 & 18.07±11.86 & \underline{10.53±10.50} \\
          & CALR~\citep{Calr} & \underline{24.23±10.77} & 34.21±9.36 & 26.27±9.79 & 19.28±10.17 & \underline{11.94±10.78} & 18.17±9.68 \\
          & FPS~\citep{Fps} & 31.72±12.60 & 29.55±9.13 & 24.95±9.47 & 20.77±9.86 & 13.74±10.75 & 10.74±10.40 \\
          & Probcover~\citep{Probcover} & \textbf{22.48±6.75} & \textbf{11.19±2.69} & \underline{17.29±4.93} & 20.12±6.98 & 25.64±9.71 & 15.23±7.61 \\
          & Typiclust~\citep{Typiclust} & 32.54±11.97 & 29.24±12.41 & 26.30±11.78 & 23.60±11.14 & 19.01±10.33 & 14.17±9.87 \\
          & CEC~\citep{Cec} & 30.76±7.78 & 27.74±7.85 & 22.31±8.61 & 18.36±9.37 & 13.20±10.40 & 11.77±9.43 \\
          & SUGFW+ (Ours) & 29.00±12.19 & \underline{16.14±12.52} & \textbf{13.68±11.17}$^*$ & \textbf{11.70±9.94}$^*$ & \textbf{8.77±8.26}$^*$ & \textbf{7.35±5.42}$^*$ \\
          \bottomrule

    \end{tabular}}%
  \label{tab:main:Liver}%
\end{table*}%

\subsection{Implementation details}
Our experiments were implemented in PyTorch on a Linux server equipped with 4 NVIDIA GeForce RTX 2080Ti GPUs. For all the four datasets, the 2D images or slices in 3D volumes were resized to \(512 \times 512\) pixels during pre-processing, and intensity values above the 99.5th percentile were clipped to suppress outliers. For the 3D datasets, we used slice-level annotation and segmentation, % processed individual slices during sample selection, 
as SAM supports only 2D inputs and the inter-slice spacing has a large variance. The training sets of Promise12, UTAH, and Liver contain 960, 5,041, and 13,792 slices, respectively, and ISIC 2018 contains 2,595 2D images. For CSAL, we set the slice-level annotation budget to the range of 3\% to 8\% for Promise12, 0.10\% to 0.75\% for UTAH, 0.05\% to 0.30\% for Liver, and 0.50\% to 3.00\% for ISIC 2018, corresponding to a maximum of 100 annotated slices per dataset. %All images were resized to 512 × 512 pixels, and intensity values above the 99.5th percentile were clipped to suppress outliers.

The ViT-Base checkpoint of the pre-trained SAM was used in the experiment.
For the surrogate uncertainty estimation with SAM, we used the ``everything'' mode with  a \(32 \times 32\) grid of 1,024 points over the image as the foreground prompt, and set $K$ to 10 following previous works~\citep{wang2019aleatoric}.  The  data augmentation used for uncertainty estimation and model fine-tuning include gamma correction, random flip, random rotation, posterization, and adjustment of contrast, sharpness  and brightness.  In the UPFT stage, we used LoRA with a rank of 32 in the frozen image encoder, and kept parameters in the uncertainty encoder, uncertainty predictor and mask decoder learnable. The model was finetuned   for 300 epochs using the AdamW optimizer, with a batch size of 16, an initial learning rate of 0.0008, a weight decay of 0.1, and \((\beta_1, \beta_2) = (0.9, 0.999)\).  $\omega$ was set to 0.8 according to  ~\cite{chen2024ma}. %, the parameters $K$ and $\omega$ were empirically set as 10 and 0.8, respectively. 
The ramp-up iteration for $\alpha_t$ was 300, and we set the two key hyper-parameter of our method as  \(\lambda = 6\) and \(\alpha_{max} = 0.02\) according to the best performance in ablation study.   % During the fine-tuning stage, the pre-trained SAM image encoder is kept largely frozen, while the trainable parameters are strictly limited to four components: the domain-specific adapter, the mask decoder, the newly introduced uncertainty encoder, and LoRA modules. Specifically, the LoRA modules are injected directly into the transformer layers of the image encoder, with the intrinsic rank set to $r=32$.

For quantitative evaluation of the segmentation results, we used the Dice Similarity Coefficient (DSC) and the 95th percentile Hausdorff Distance (HD95). For each casse of the 3D datasets (Promise12, UTAH, and Liver), the slice-level segmentation results were stacked into a 3D volume, and   both metrics were calculated in the 3D space.

\begin{table*}[t!]
  \centering
  \caption{Quantitative comparison of different CSAL methods on the ISIC 2018 dataset. The best results are in bold, and the second-best are underlined. $*$ indicates that the p-value $<$ 0.05 in paired t-test compared to the second-best results.}
  \resizebox{\textwidth}{!}{
    \begin{tabular}{p{1cm}|c|ccccccc}

    \toprule
    \multirow{2}{*}{Metric} & \multirow{2}{*}{Method} & \multicolumn{6}{c}{Annotation Budget} \\
     & & 0.5 \%    & 1.0\%    & 1.5\%    & 2.0\%    & 2.5\%    & 3.0\% \\
    \midrule
    \multirow{8}[2]{*}{\makecell{DSC \\ (\%) $\uparrow$}} & Random & 83.67±15.75 & 83.99±15.91 & 84.08±15.18 & 84.46±16.31 & 86.23±16.18 & 86.42±16.06 \\
          & ALPS~\citep{ALPS} & 82.60±21.79 & 82.52±15.00 & \underline{86.56±17.02} & 87.21±13.08 & \underline{87.67±14.23} & \underline{88.98±13.07} \\
          & CALR~\citep{Calr} & \underline{84.00±15.00} & \underline{85.26±14.55} & 85.23±17.16 & 86.74±12.59 & 87.18±12.56 & 88.45±14.38 \\
          & FPS~\citep{Fps} & 82.91±15.19 & 83.86±17.82 & 85.10±14.50 & 86.14±12.37 & 87.19±13.91 & 87.50±11.74 \\
          & Probcover~\citep{Probcover} & 79.95±15.14 & 80.65±18.31 & 84.47±17.82 & 84.57±18.87 & 84.92±17.33 & 87.22±15.86 \\
          & Typiclust~\citep{Typiclust} & 61.50±17.87 & 83.43±14.96 & 84.22±19.75 & \underline{87.25±12.72} & 87.55±13.56 & 88.84±13.12 \\
          & CEC~\citep{Cec} & 80.08±21.90 & 84.18±16.35 & 84.57±12.53 & 86.36±13.65 & 86.71±15.74 & 87.32±16.01 \\
          & SUGFW+ (Ours) & \textbf{86.36±13.50}$^*$ & \textbf{87.08±13.74}$^*$ & \textbf{87.59±14.22}$^*$ & \textbf{88.69±12.50}$^*$ & \textbf{88.86±15.48}$^*$ & \textbf{89.75±12.44}$^*$ \\

    \midrule

    \multirow{8}[2]{*}{\makecell{HD95 \\ (pixel) $\downarrow$}} & Random & \underline{18.59±38.89} & 25.02±32.69 & 19.44±40.71 & 31.58±37.56 & 14.91±37.81 & 19.49±38.47 \\
    & ALPS~\citep{ALPS} & 29.22±50.64 & 18.85±39.28 & 21.80±44.81 & 20.73±40.50 & 17.26±36.76 & 17.39±37.11 \\
    & CALR~\citep{Calr} & 40.47±47.46 & 23.34±38.26 & 22.52±34.77 & 21.10±35.75 & 18.19±38.23 & 15.78±30.36 \\
    & FPS~\citep{Fps} & 34.35±49.05 & 27.41±40.43 & 19.91±34.89 & 21.96±39.82 & 20.46±35.00 & 16.77±38.15 \\
    & Probcover~\citep{Probcover} & 23.33±44.34 & 27.67±37.88 & 22.09±40.33 & 18.67±33.17 & 18.28±40.98 & 20.65±42.60 \\
    & Typiclust~\citep{Typiclust} & 22.90±39.57 & 25.68±32.25 & 18.18±40.50 & 16.46±36.47 & 14.99±36.45 & 20.17±38.00 \\
    & CEC~\citep{Cec} & 22.41±41.86 & \underline{18.13±36.45} & \underline{17.32±39.89} & \underline{16.39±34.47} & \underline{14.21±32.77} & \underline{15.34±32.66} \\
    & SUGFW+ (Ours) & \textbf{17.59±36.14}$^*$ & \textbf{17.31±35.19}$^*$ & \textbf{16.25±31.73}$^*$ & \textbf{16.12±34.06} & \textbf{14.07±35.31} & \textbf{13.32±33.96}$^*$ \\

    \bottomrule
     \end{tabular}}%
  \label{tab:main:ISIC:Dice}%
\end{table*}%

\subsection{Comparison with state-of-the-art methods}
We compared our method with seven state-of-the-art CSAL methods.
  1) \textbf{Random}: randomly selecting samples;
  2) \textbf{ALPS}~\citep{ALPS}: choosing the sample closest to the cluster center;
  3) \textbf{CALR}~\citep{Calr}: performing hierarchical clustering and selecting the cluster centers; % to ensure maximum representativeness of the data distribution;
  4) \textbf{FPS}~\citep{Fps}: combining clustering with Farthest Point Sampling  on features;
  5) \textbf{ProbCover}~\citep{Probcover}: constructing a graph on features and selecting a set of samples that maximizes the probability coverage within a specified distance threshold;
  6) \textbf{TypiClust}~\citep{Typiclust}: partitioning the feature space into clusters and selecting the most typical sample (the one with the highest local density) from each cluster;
  7) \textbf{CEC}~\citep{Cec}: utilizing calibrated entropy and neighbor-aware uncertainty to select the most informative samples.
 For a fair comparison, all these methods used the same SAM ViT-Base encoder for feature extraction, and  global average pooling over encoder's output feature map is used to obtain image-level feature. Note that in our preliminary work~\citep{ma2025sugfw}, the different sample selection methods were compared with the same U-Net model for training. In this work, we compare the sample selection methods with our UPFT-based SAM as the segmentation model. The SUGFW+ in this work is also compared with its preliminary version SUGFW in Sec.~\ref{sec:ablation}. %  We apply our uncertainty-prompted fine-tuning strategy as the downstream segmentation model for all compared methods, so that performance differences reflect the sample selection strategy   rather than the training procedure. The seven baselines are summarized below:

\begin{figure*}[t!]
    \centerline{\includegraphics[width=18cm]{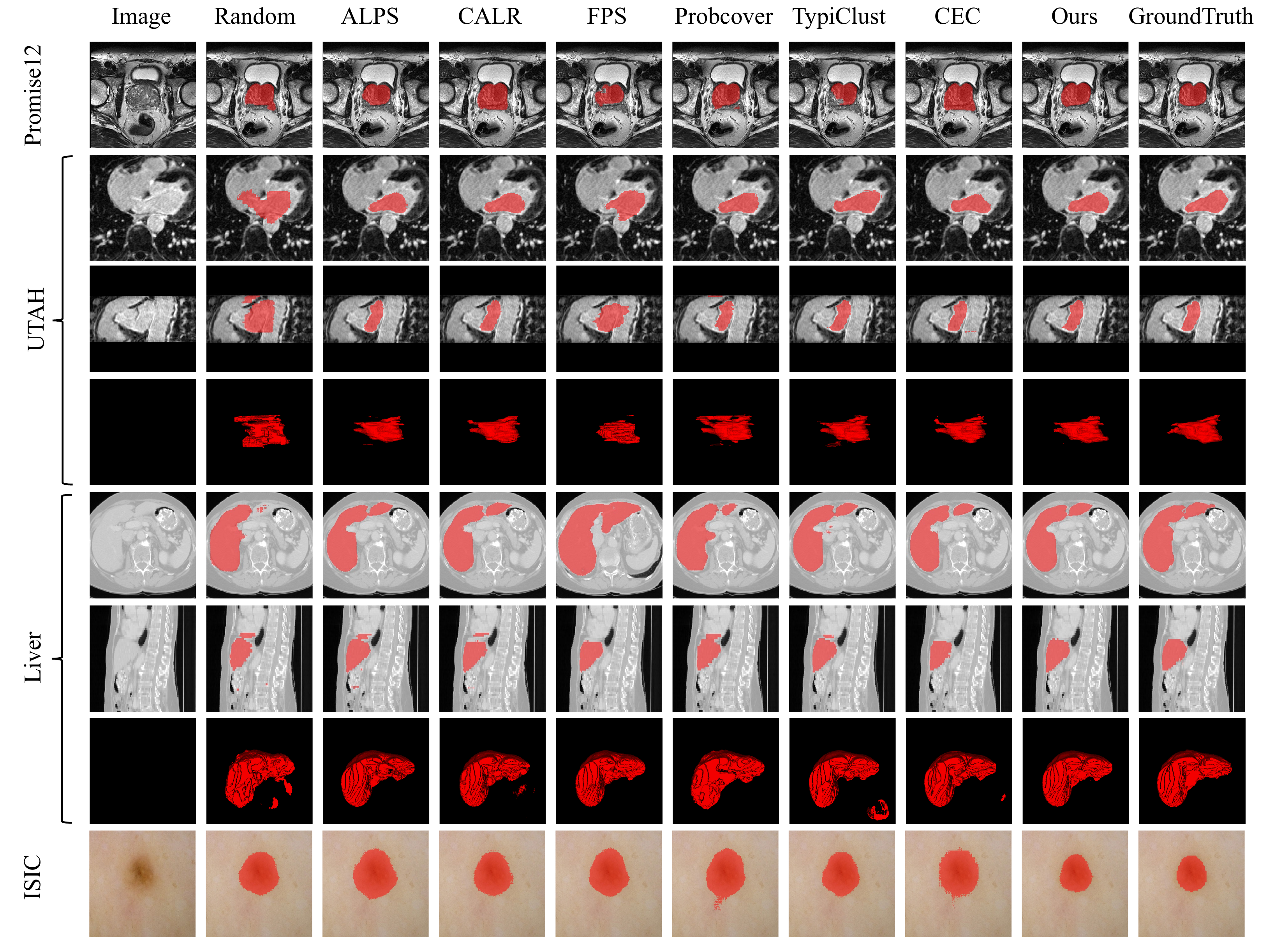}}
    \caption{Visual comparison of our method and other CSAL methods. All datasets are presented in axial views; additionally, the sagittal views and 3D reconstructions are provided for the left atrium and liver segmentations.}
    \label{fig: visualization}
\end{figure*}

\subsubsection{Result for prostate segmentation}
Table~\ref{tab:main:Promise12} shows performance of different CSAL methods on the Promise12 dataset for prostate segmentation under a range of  annotation budgets. At a 3\% annotation budget, our method achieves a DSC of 86.38\%, outperforming the best existing method FPS (84.24\%) by a clear margin. As the annotation budget increases to 8\%, our method reaches a DSC of 89.25\%, outperforming the second-best method Probcover (88.88\%). %In terms of HD95, our method achieves the best results across all evaluated annotation budgets, notably recording the lowest HD95 of 1.35 $mm$ at the 8\% budget. 
Compared to the top existing methods ProbCover~\citep{Probcover} and FPS~\citep{Fps}, %which also show competitive performance in higher budgets, 
our approach obtains more consistent improvements across both DSC and HD95 metrics, especially in the extremely low annotation budgets (3\%--5\%). Visual comparison of the segmentaiton results in Fig.~\ref{fig: visualization} further demonstrates this superiority. In the Promise12 row of Fig.~\ref{fig: visualization}, our method exhibits the highest overlap and closest fit to the ground truth contour.

\subsubsection{Result for left atrium segmentation}
For left atrium segmentation on the UTAH dataset, the results of CSAL methods under six different annotation budgets ranging from 0.10\% to 0.75\% are shown in Table~\ref{tab:main:UTAH}. At the 0.10\% budget, our method achieves a DSC of 78.70\%, which is significantly higher than Random  baseline (70.96\%) and top existing methods including TypiClust~\citep{Typiclust} (74.03\%) and ALPS~\citep{ALPS} (73.80\%). This advantage is maintained as the budget increases, with our method reaching a DSC of 89.44\% at only a 0.75\% annotation budget. For HD95, our method consistently shows superior results across all budget levels, notably achieving the lowest HD95 of 7.30 $mm$ at the 0.75\% budget. %While methods such as ALPS~\citep{ALPS} and CEC~\citep{Cec} show competitive performance in specific scenarios, our method provides more robust and consistent improvements as the annotation budget scales. 
As shown in the visual comparison of Fig.~\ref{fig: visualization}, models trained with the existing CSAL methods often suffer from erroneous over-segmentation or fail to capture the complete structure under extremely limited supervision. In contrast, the  model trained with our sample selection strategy achieves a more precise and integrated segmentation of the target region, closely aligning with the ground truth.

\subsubsection{Result for liver segmentation}
For liver segmentation, as reported in Table~\ref{tab:main:Liver}, our proposed SUGFW+ consistently outperforms state-of-the-art CSAL methods different low annotation budgets.  At an ultra-low budget of 0.05\%, our method achieves a DSC of 82.27\%, markedly surpassing the Random baseline (68.99\%) and the second-best method TypiClust (77.07\%). As the annotation budget scales to 0.30\%, our approach maintains its dominance, reaching a DSC of 93.11\%. % and effectively narrowing the performance gap typically associated with limited supervision.
Regarding the HD95 metric, our method also outperformed the best existing method Probcover in four out of the six annotation budget settings. Though Probcover shows competitive HD95 results at the  0.05\% and 0.06\% annotation budgets, the HD95 value fluctuates with an increase of annotation ratio, e.g., HD95 increases from 11.19~mm to 25.64~mm when the annotation ratio increases from 0.06\% to 0.20\%. In contrast, our method obtains a consistent decline of HD95 with increase of annotation budgets, showing its stability over  Probcover.  %demonstrates superior boundary precision and stability as the labeling proportion increases. Specifically, from a 0.08\% budget onwards, SUGFW+ achieves the lowest HD95 values, significantly outperforming methods that exhibit high variance or erratic performance in sparse data regimes, such as ALPS and CEC. While Probcover shows competitive HD95 results at the initial 0.05\% and 0.06\% stages, its DSC performance lags significantly, highlighting our method's ability to balance regional overlap and boundary accuracy. 
The visualization results in Fig.~\ref{fig: visualization} further confirm that our method provides more contiguous and anatomically precise liver contours, whereas alternative methods often suffer from over-segmentation or fail to capture the complete morphology of the liver under the low annotation budgets in CSAL. % extreme labeling constraints.

\begin{table*}[t!]
    \centering
    \caption{Ablation study on the sample selection strategy SUGFW  on Promise12 and ISIC 2018 datasets. $U_{PFUC}$ means unceratinty estiamtion based on PFUC, and $F_{PGDR}$
    means image-level feature extraction based on PGDR.  UPFT of SAM is used for model training after querying. }
    \resizebox{\textwidth}{!}{
      \begin{tabular}{c|ccc|ccc|ccc}
      \toprule
      \multirow{2}{*}{Method} & \multirow{2}{*}{$U_{PFUC}$} & \multirow{2}{*}{$F_{PGDR}$} & \multirow{2}{*}{GSCU}& \multicolumn{3}{c|}{Promise12} & \multicolumn{3}{c}{ISIC 2018} \\
       & & & & 3.0\% & 4.0\% & 5.0\% & 0.5\% & 1.0\% & 1.5\% \\
      \midrule
    Random  & $\times$ & $\times$ & $\times$
            & 84.03±5.87 & 85.72±4.86 & 85.74±4.33
            & 83.67±15.75 & 83.99±15.91 & 84.08±15.18 \\
   Un only  & \checkmark & $\times$ & $\times$
            & 85.73±5.05& 86.60±4.07& 87.71±4.00
            & 85.94±12.40& 86.09±13.31& 86.94±13.56\\
w/o GSCU    & \checkmark & \checkmark & $\times$
            & 81.22±6.18& 85.32±6.78& 87.55±5.24
            & 82.46±17.38& 83.72±14.85& 86.26±12.26\\
  w/o PGDR  & \checkmark & $\times$ & \checkmark
            & 85.13±4.66& 85.30±5.25& 87.49±5.61
            & 83.85±16.23& 86.19±13.51& 86.58±11.47\\
 SUGFW      & \checkmark & \checkmark & \checkmark
            & \textbf{86.38±5.60}& \textbf{87.35±3.90}& \textbf{87.86±5.86}
            & \textbf{86.36±13.50}& \textbf{87.08±13.74}& \textbf{87.59±14.22}\\
      \bottomrule
      \end{tabular}
    }
    \label{tab:ablation::sample_selection}
  \end{table*}

 \begin{table*}[t!]
    \centering
    \caption{Ablation study on model training methods on Promise12 and ISIC 2018 datasets. The query samples are selected by our SUGFW strategy. }
    \resizebox{\textwidth}{!}{
    
      \begin{tabular}{C{2cm}|C{2cm}|ccc|ccc}
      \toprule
        \multirow{2}{*}{Method} & \multirow{2}{*}{\makecell{Trainable \\ Params (M)}} & \multicolumn{3}{c|}{Promise12} & \multicolumn{3}{c}{ISIC 2018} \\
      & & 3.0\% & 4.0\% & 5.0\% & 0.5\% & 1.0\% & 1.5\% \\
      \midrule
            SwinUNet & 62.19 & 60.59±2.01& 64.47±5.43& 72.63±2.86& 57.85±27.76& 59.26±27.40& 61.44±26.10\\
            nnUNet     & 32.42 & 63.07±10.51& 68.41±7.31& 76.05±5.03& 70.87±25.80& 74.03±26.12& 76.96±23.83\\
            DINOUNet & 11.18 & 85.72±4.60& 86.34±4.08& 86.75±3.95& 85.11±15.75& 85.94±13.49& 86.77±13.37\\
            Ours w/o Un & 28.67 & 85.01±5.67& 86.72±4.23& 87.52±5.38& 86.02±13.80& 86.88±14.93& 87.02±13.54\\
            Ours                      & 29.86 & \textbf{86.38±5.60}& \textbf{87.35±3.90}& \textbf{87.86±5.86}& \textbf{86.36±13.50}& \textbf{87.08±13.74}& \textbf{87.59±14.22}\\
      \bottomrule
      \end{tabular}
      }
    \label{tab:ablation::model_training}
  \end{table*}

\subsubsection{Result for skin lesion segmentation}
Table~\ref{tab:main:ISIC:Dice} shows the quantitative evaluation results for 2D skin lesion segmentation on the ISIC dataset. %, our method consistently maintains its performance edge under restricted annotation budgets. As demonstrated in , 
At a minimal 0.5\% budget, our proposed SUGFW+ achieves a DSC of 86.36\%, significantly outperforming the Random baseline (83.67\%) and the leading existing CSAL methods such as CALR~\citep{Calr} (84.00\%) and FPS~\citep{Fps} (82.91\%). This superiority is sustained as the annotation budget scales, with our method reaching a DSC of 89.75\% at the 3.0\% budget level. Notably, while ALPS~\citep{ALPS} outperformed the other existing CSAL methods under the 2.5\% and 3.0\% annotation budgets, its performance is lower than the random baseline when the annotation budget decreases to 0.5\% and 1.0\%. 
% established methods like TypiClust~\citep{Typiclust} and ALPS~\citep{ALPS} show considerable recovery and competitive results as the budget increases toward 3\%, they exhibit higher instability (e.g., lower DSC at 0.50\%) compared to our approach. 
In contrast, our method consistently obtains the highest DSC and lowest HD95 values % more reliable and consistent segmentation accuracy 
across the entire spectrum of low-budget scenarios. The visualization results for ISIC at the bottom of Fig.~\ref{fig: visualization} further confirm that our method excels at capturing the irregular boundaries and varied textures of skin lesions under a low annotation budget. %, whereas comparative techniques often produce results with lower boundary fidelity or fail to distinguish the lesion from complex background patterns under sparse supervision.

\subsection{Ablation study}\label{sec:ablation}

To evaluate the contribution of each module in our SUGFW+ framework that combines SUGFW and UPFT, we conducted ablation studies covering both sample selection and model training on the Promise12 and ISIC datasets at different low annotation budgets. In addition, we used the Promise12 dataset to analyze the effect of key hyper-parameters ($\lambda$ and $\alpha_{max}$) of our method. 

\subsubsection{Effectiveness of SUGFW for sample selection}
  To investigate the effectiveness of our querying strategy SUGFW, we compare it  against four variants in Table~\ref{tab:ablation::sample_selection}: 1) \textbf{Random Selection}: sampling entirely at random without any active learning criterion; 2) \textbf{Uncertainty Only}: selecting samples with the uncertainty $u_i$ obtained by PFUC across the dataset; (3) \textbf{w/o GSCU}: omitting the greedy diversity constraint by randomly selecting one sample from each cluster in the feature space of $f_i$ defined in Eq.~\eqref{eq:global_feature}; and 4) \textbf{w/o PGDR}: replacing the uncertainty-weighted feature fusion in Eq.~\eqref{eq:global_feature} with conventional global average pooling to obtain image-level feature for clustering. For each querying method, the queried samples are used to train the segmentation model based on UPFT of SAM.
  %The quantitative results highlight the necessity of balancing uncertainty and diversity with dedicated modules. 
 
  As shown in Table~\ref{tab:ablation::sample_selection}, the Random baseline yields the lowest performance across all budgets, while adding the SAM-derived uncertainty
  criterion improves the performance consistently. Compared to our selection strategy, removing GSCU at the lowest annotation  budget causes a significant DSC drop  from 86.38\% to 81.22\% on Promise12, and from 86.36\% to
  82.46\% on ISIC. %This indicates that relying solely on generic clustering without a greedy constraint leads to the selection of redundant samples, failing to provide sufficient informative variety. 
  Similarly, the removal of PGDR causes
  DSC to degrade to 85.13\% on Promise12 (annotation ratio 3.0\%) and 83.85\% on ISIC (annotation ratio 0.5\%), respectively. This demonstrates that conventional global average pooling is inadequate, and the uncertainty-guided  feature fusion in PGDR is more effective for image feature representation. Ultimately, combining all the modules in our sample selection method lead to the best performance on both datasets under the different annotation ratios. % while individual components can bring partial improvements, their synergistic combination in the full SUGFW+ framework consistently yields the most effective queries.

  \subsubsection{Effectiveness of our UPFT for model training}
  To demonstrate the superiority our UPFT-based SAM adaptation for training the segmentation model after annotation, we compare it with three state-of-the-art segmentation models: 1) \textbf{SwinUNet}, a Transformer-based U-Net variant with hierarchical shifted windows~\citep{SwinUNet}; 2) \textbf{nnUNet}, a self-configuring convolutional framework that adapts preprocessing and architecture to dataset characteristics~\citep{nnUNet}; and 3) \textbf{DINOUNet} that utilizes frozen DINOv3 encoder coupled with a U-Net decoder for  segmentation~\citep{DINOUNet}. In addition, we consider a variant of our method by  removing the uncertainty prompt module (Ours w/o Uncertainty), i.e., just using SAM encoder with LoRA and decoder for fine-tuning.  For a fair comparison, all these models were trained with the same set of query samples obtained by SUGFW. 
  
 Table~\ref{tab:ablation::sample_selection} shows that the conventional medical image segmentation models have a low performance under the low annotation budget, even the query samples are provided by our method. Notably, SwinUNet~\citep{SwinUNet}, despite the larger model size
  (62.19M) than the other models, achieves DSC of only 60.59\% on Promise12 and 57.85\% on ISIC at the  annotation ratios of 3.0\% and 0.5\%, respectively. Similarly, nnUNet~\citep{nnUNet} with a model size of 32.42M also struggles in these extreme low-budget regimes. %This proves that in cold-start   scenarios, informative samples alone are insufficient if the network lacks powerful pre-trained priors to process them.
  In contrast, the three foundation model-based methods outperformed the traditional models trained from scratch, showing higher performance under the limited annotated samples with even fewer trainable
  parameters. Importantly, our  framework consistently outperforms DINOUNet and Ours w/o Uncertainty across all budgets. On the Promise12 dataset with a budget of 3.0\%, the use of uncertainty prompt improves the DSC  from 85.01\% to 86.38\% when fine-tuning SAM, and on the ISIC 2018 dataset with annotation budget of 0.5\%, it also improves the DSC from 86.02\% to 86.36\%, while maintaining a highly comparable parameter overhead. These results demonstrates the effectiveness of our UPFT-based SAM in learning from limited labeled data in CSAL. %that explicitly guiding the SAM backbone to focus on uncertain regions during the fine-tuning stage, rather than treating   these queried samples as standard training data, is crucial for maximizing their informational yield without requiring a substantial increase in model capacity.

\subsubsection{Hyper-parameter study}
Our method has two main hyper-parameters: $\lambda$ in Eq.~\eqref{eq:global_feature} for scaling the patch-wise uncertainty during fusion, and $\alpha_{max}$ that controls the weight of uncertainty prediction loss in Eq.~\eqref{eq:total_loss}. % We then investigate the effectiveness of the $\lambda$ and the $\alpha_{max}$.

\begin{figure}%[htbp]
    \centerline{\includegraphics[width=9cm]{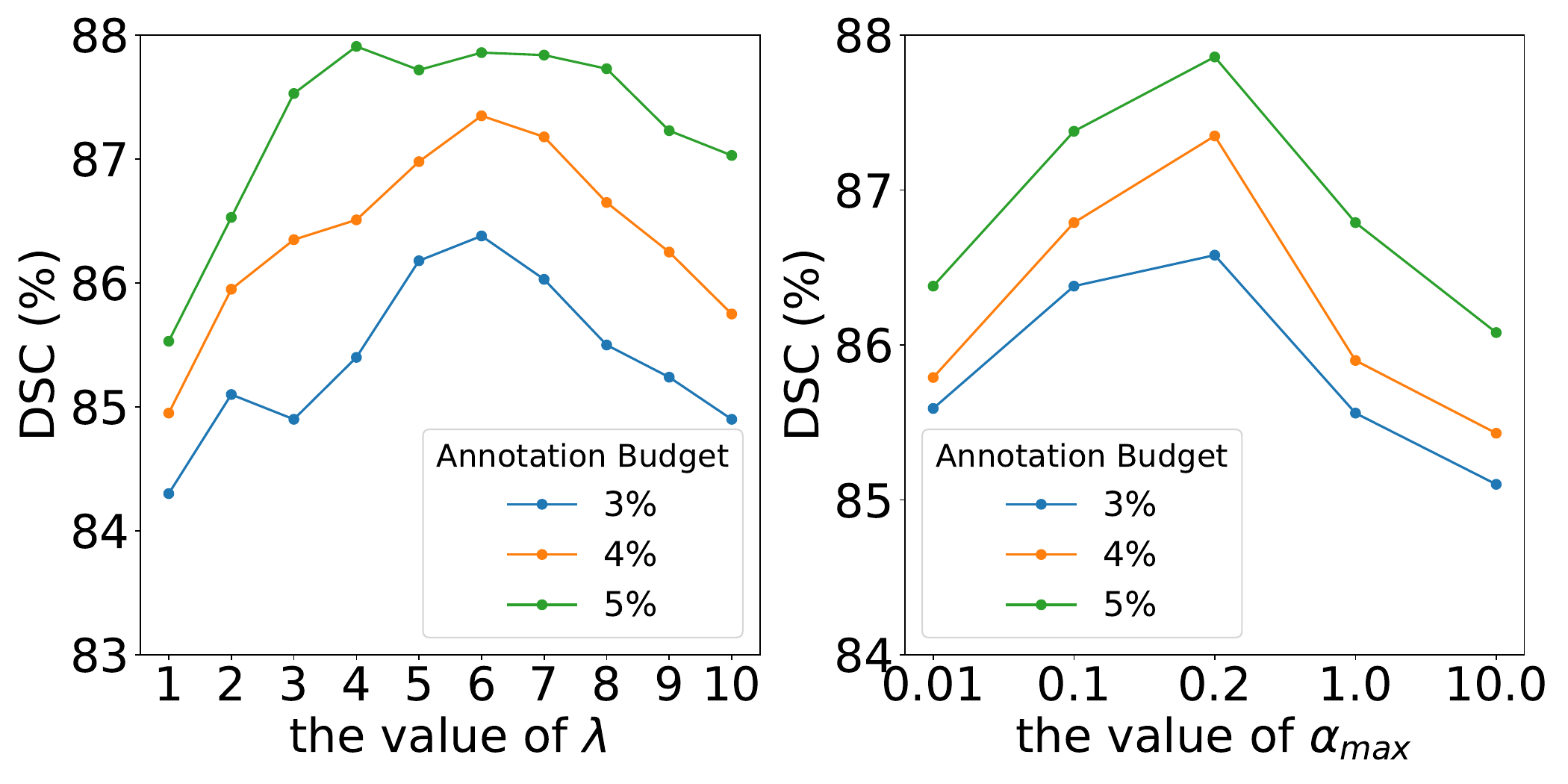}}
    \caption{Hyper-parameter analysis of $\lambda$ and $\alpha_{max}$ on the Promise12 dataset.}
    \label{fig: param1}
\end{figure}

As illustrated in Fig.~\ref{fig: param1} (left), we evaluate the impact of $\lambda$ across an extended range from 1 to 10 on the Promise12 dataset. The segmentation performance shows a clear ascending trend as $\lambda$ increases from 1.0 to 6.0, with the highest  DSC scores at $\lambda = 6.0$ across the three annotation budgets, suggesting that moderately amplifying the contribution of surrogate uncertainty during patch-level feature aggregation is helpful for more effective image-level representation. However, when $\lambda$ becomes even larger, the performance begins to degrade, which is likely due to that over-amplification of the surrogate uncertainty will ignore the common regions in the image, reducing the representativeness of image-level features. Therefore, we set   $\lambda = 6.0$ in the experiments. 

Furthermore, as shown in Fig.~\ref{fig: param1} (right), we vary $\alpha_{\max}$ logarithmically from 0.01 to 10.0. The model performance is relatively sub-optimal when the weight is too small ($\alpha_{\max} = 0.01$). %, indicating that insufficient attention is directed toward the hard, uncertain regions. 
As $\alpha_{\max}$ increases to 0.20, the framework achieves the optimal segmentation performance across all annotation budgets. %, securing the highest DSC scores (86.58\%, 87.35\%, and 87.86\% for 3\%, 4\%, and 5\% budgets, respectively). 
In addition, setting $\alpha_{\max}$ too high (e.g., 1.0 or 10.0) causes a noticeable performance drop, which is due to the relatively decreased weight of the supervised segmentation loss. %This indicates that because the uncertainty encoder has not yet fully converged during the early training stages, forcing a strong alignment via excessive weighting introduces noisy and unreliable guidance. 
Therefore, we set $\alpha_{\max} = 0.20$ to obtain an optimal balance between the loss terms.

\begin{figure}[htbp]
    \centerline{\includegraphics[width=9cm]{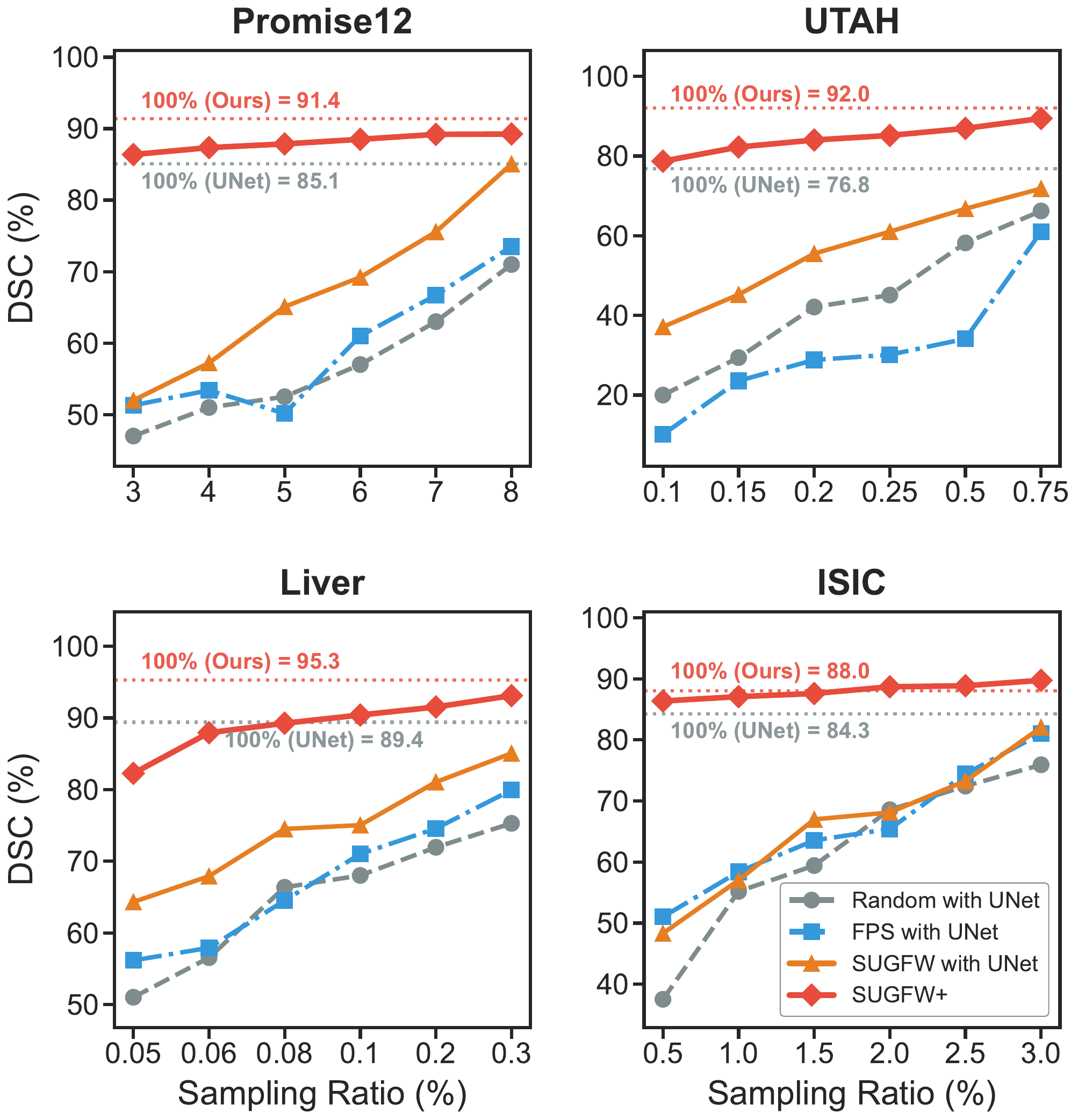}}
    \caption{Extended analysis of SUGFW and SUGFW+.}
    \label{fig:extended_experiments}
\end{figure}

%\textbf{Performance under Extremely Low Annotation Budgets.} Figure~\ref{fig:extended_experiments} presents the segmentation performance on the Promise12, UTAH, Liver, and ISIC datasets under extremely low sampling ratios. Across all datasets, the proposed SUGFW+ framework consistently outperforms the random selection baseline, the FPS strategy, and our preliminary SUGFW. The performance gap is particularly striking at the strictest initial budgets. For instance, at a mere 3\% sampling ratio on the Promise12 dataset, traditional random selection and FPS achieve DSC of only 47.13\% and 51.36\%, respectively. In contrast, SUGFW+ immediately secures a high DSC of 86.38\%. Similar patterns are observed on the UTAH dataset at 0.10\% annotation (78.74\% for SUGFW+ vs. 10.15\% for FPS) and the Liver dataset at 0.05\% annotation (82.27\% for SUGFW+ vs. 51.02\% for random selection). This massive early-stage performance leap effectively solves the "cold start" challenge, proving that our framework not only extracts the most critical discriminative information when labeled data is extremely scarce, but also effectively adapts the foundation model to maximize the informational yield of these queried samples.

%\textbf{Superiority over Preliminary SUGFW and Full-Supervision Approximations.} The results further underscore the significant advancement of the SUGFW+ over its predecessor. 

\subsection{Comparison with Full Supervision and SUGFW}
To further demonstrate the robustness and superiority of our SUGFW+ (using UPFT of SAM for model training), we compared it with  the preliminary version SUGFW (using UNet for model training) and the full annotation upper bound on the four datasets, and the results are shown in Fig.~\ref{fig:extended_experiments}. 
With the UNet backbone for model training, SUFGW outperformed random selection and FPS on the four datasets under different annotation budgets. On the Promise12 dataset, with an annotation ratio of 8\%, SUGFW obtained a performance close to that of  full annotation. It can be observed that replacing the UNet backbone with our UPFT-based SAM (i.e., SUGFW+) substantially improved the performance on all the four datasets. Especially on the Promise12 dataset, the latter improved the DSC score from 85.1\% to 91.4\% for the fully annotated supervised training. Our SUGFW+ with only 3\% annotation even outperformed the fully supervised UNet, and improved the average DSC score by around 35 percentage  points compared with SUGFW. On the UTAH and ISIC 2018 datasets, our SUGFW+ also outperformed the fully supervised UNet while reducing the annotation cost to 0.1\% and 0.5\% respectively, and largely outperformed FPS and SUGFW using UNet. The results demonstrate that the segmentation model plays an important role in effective learning with the small number of annotated samples selected by CSAL, and the combination of SUGFW and our UPFT-based SAM can effectively obtain high-performance segmentation models under an extrmely low annotation budget.  
% While the preliminary SUGFW relies on standard UNet training and shows only marginal improvements over FPS at early stages (e.g., achieving 52.07\% on Promise12 at a 3\% budget), it still struggles to provide clinically satisfactory segmentation in extreme low-data regimes. By integrating our active selection strategy with the foundation model adaptation, SUGFW+ bridges this gap completely. Remarkably, as the budget slightly increases, SUGFW+ rapidly converges toward, and sometimes even exceeds, the upper bound performance achieved by training on 100\% of the data. For example, on Promise12 and Liver datasets, SUGFW+ achieves 89.25\% and 93.11\%, closely approaching their full-supervision upper bounds of 91.39\% and 95.30\%. Most notably, on the ISIC dataset, SUGFW+ attains a DSC of 89.75\% at a 3.00\% sampling ratio, surpassing the 100\% full-supervision baseline of 88.01\%. These findings indicate that our SUGFW+ strategy not only minimizes the annotation burden but also effectively filters out noisy or low-quality instances, yielding a more robust and generalizable model than conventional full-set training.

\section{Discussion}
The proposed framework demonstrates profound efficacy in CSAL for medical image segmentation, and consistently establishes a new state-of-the-art across diverse medical datasets. The experimental results show that both sample selection strategy and the segmentation model are critical for obtaining a high performance. Traditional CSAL methods often focus on the sample selection strategy and use standard segmentation models like UNet for training. Though the strategical sample selection method is effective, the performance is still limited by the segmentation model and the training strategy. Results in Fig.~\ref{fig:extended_experiments} show that the UPFT-based SAM could improve the DSC of fully supervised learning by 5.7\%, 11.2\%, 5.9\% and 3.7\% percentage points on the Promise12, UTAH, MSD Liver and ISIC 2018 datasets, respectively, demonstrating its superiority over traditional UNet models. Table~\ref{tab:ablation::model_training} and Fig.~\ref{fig:extended_experiments} also demonstrate that UPFT-based SAM outperformed other segmentation models in the CSAL setting, showing the importance of our UPFT-based SAM in learning from small annotated data in CSAL.

% Specifically, on the Promise12, UTAH, and Liver datasets, SUGFW+ achieves performance comparable to the 100\% full-supervision upper bound while utilizing only few labeled samples (e.g., 0.75\% for UTAH). Most notably, the significant performance leap from 53.58\% (SUGFW) to 86.38\% (SUGFW+) at a 3\% annotation budget underscores our framework's superior adaptation to extremely scarce data scenarios. This ``performance leap'' effectively addresses the ``cold-start'' challenge that often hinders traditional UNet-based active learning. Furthermore, the observation that SUGFW+ can even outperform full supervision on the ISIC dataset (89.75\% vs. 88.01\%) suggests that our selection mechanism acts as a robust filter against label noise.

Our sample selection strategy SUGFW is also critical in performance improvement even with the strong UPFT-based SAM model. Table~\ref{tab:main:Promise12} shows that ALPS, CLAR and CEC performed worse than Random selection with an annotation budget of 3.0\% to 5.0\%, demonstrating the reduced effectiveness of existing methods under extremely low annotation budget in CSAL. Our SUFGW is the only one  method that outperformed Random selection on all the experimental datasets under all the considered low annotation budgets, indicating its high  robustness and generalizability. Its success can be attributed to the core methodological innovations: Firstly, our PFUC strategy leveraged the foundation model SAM for feature extraction, and PGDR weights patch-level features based on uncertainty, resulting in a global representation that is far more discriminative. %This ensures that the subsequent clustering step groups samples based on meaningful semantic differences rather than background noise. Secondly, the robust sample selection strategy. 
Secondly, the GSCU strategy effectively balances representativeness and diversity. While clustering ensures geometric coverage of the feature space, the greedy mechanism ensures coverage of the uncertainty spectrum. This dual approach prevents the selection of redundant easy samples or outliers, providing the model with a training subset that optimally mimics the distribution of the full dataset. %Thirdly, the uncertainty-prompted SAM fine-tuning strategy. This strategy effectively directs the model's focus toward reliable regions, thereby facilitating efficient feature learning. Furthermore, a dedicated predictor is employed to learn the uncertainty embedding, providing an expedited mechanism for uncertainty estimation during inference and significantly reducing computational overhead.

While our framework demonstrates effective performance in CSAL, we identify a couple of limitations that can be addressed in the future. First, the efficiency of the sample selection phase could be further improved. Our  uncertainty estimation in PFUC relies on $K$-time augmentation and the ``everything mode'' of SAM. % to ensure maximum precision in highlighting informative regions. 
While such a configuration remains computationally manageable for common medical datasets and does not impact real-time clinical deployment, it introduces a potential overhead when scaling to massive, petabyte-scale unlabeled repositories. Therefore, more efficient uncertainty estimation methods could be investigated %To broaden the applicability of SUGFW+ across even more resource-constrained environments, our future work will focus on developing accelerated approximation techniques for uncertainty estimation, aiming 
to minimize time costs while maintaining high-fidelity sample selection. Second, this work has only considered 2D images or slices for annotation and model training due to the 2D structure of SAM. It is of interest to extend our framework to 3D networks by using a corresponding 3D version of SAM in the future.

\section{Conclusion}
We presented SUGFW+, a novel cold start active learning framework that effectively adapts SAM for medical image segmentation under very limited annotation budget. By introducing the Patch-level Feature and Uncertainty Calculation (PFUC) via SAM and Patch-based Global Distinct Representation (PGDR), we obtain uncertainty-aware distinctive feature representations of the unannotated images, and then the Greedy Selection with Cluster and Uncertainty (GSCU) strategy encourages sample representativeness  in the feature and uncertainty space simultaneously. Furthermore, the integration of Uncertainty-Prompted Fine-tuning (UPFT) of SAM ensures that the model learns robustly from the limited annotated samples. Extensive evaluations on four datasets demonstrated the superiority of our framework over existing CSAL methods, and it offers a promising solution for drastically reducing the annotation burden while keeping high performance in medical image segmentation.

\section*{Acknowledgment}
This work was supported by the National Natural Science Foundation
of China (62271115), and the Natural Science Foundation of Sichuan Province (2025ZNSFSC0455, 2026NSFSCZY0030).

\bibliographystyle{model2-names.bst}\biboptions{authoryear}
\bibliography{refs}

\end{document}